# Connecting the Persian-speaking World through Transliteration

Rayyan Merchant1, Akhilesh Kakolu Ramarao2, and Kevin Tang1, 2

1Department of Linguistics, University of Florida, Gainesville, Florida, 32611-5454, United States of America
2Department of English Language and Linguistics, Institute of English and American Studies, Heinrich-Heine-University, Düsseldorf, 40225, Germany

February 26, 2025

## Abstract

Despite speaking mutually intelligible varieties of the same language, speakers of Tajik Persian, written in a modified Cyrillic alphabet, cannot read Iranian and Afghan texts written in the Perso-Arabic script. As the vast majority of Persian text on the Internet is written in Perso-Arabic, monolingual Tajik speakers are unable to interface with the Internet in any meaningful way. Due to overwhelming similarity between the formal registers of these dialects and the scarcity of Tajik-Farsi parallel data, machine transliteration has been proposed as more a practical and appropriate solution than machine translation. This paper presents a transformer-based G2P approach to Tajik-Farsi transliteration, achieving chrF++ scores of 58.70 (Farsi to Tajik) and 74.20 (Tajik to Farsi) on novel digraphic datasets, setting a comparable baseline metric for future work. Our results also demonstrate the non-trivial difficulty of this task in both directions. We also provide an overview of the differences between the two scripts and the challenges they present, so as to aid future efforts in Tajik-Farsi transliteration.

## 1 Introduction

Tajik Persian (henceforth, Tajik) is the formal variety of Modern Persian spoken in Tajikistan. As such, it retains an extremely high level of mutual intelligibility with formal Persian as spoken in Iran and Afghanistan (henceforth referred to as Farsi). Unlike these two countries which use the centuries-old Perso-Arabic script, Tajikistan uses the relatively new Tajik-Cyrillic script due to Tajikistan's Soviet heritage (Perry 2005).

While proposals have been made to shift the script back to Perso-Arabic, any significant shift will likely not occur in the near future, with Tajikistan's former Minister of Culture stating in 2008 that "...some 90-95% of Tajikistan's population is not familiar with Arabic script..."

(Ghufronov 2008). As a result, the vast majority of the 10 million Persian speakers in Tajikistan are unable to read written Persian media produced by the 100 million Persian speakers in Iran and Afghanistan. This restriction extends to the Internet, where Farsi dominates. As of September 2023, the Tajik Wikipedia had 269,857 articles and 10.5 million words across all content pages compared to the Farsi Wikipedia's 5.5 million articles and 194 million words (Wikimedia Foundation 2023b). Moreover, despite Afghan Persian (known as Dari) being similarly divergent, the Farsi Wikipedia is the most popular Wikipedia in Afghanistan and there currently exists no Dari Wikipedia (Wikimedia Foundation 2023a).

These two scripts are highly incongruous (Perry 2005). The Perso-Arabic script, as an impure abjad, often omits vowels, and those that are written are ambiguous. Meanwhile, the Tajik-Cyrillic script, as an alphabet, writes out all vowels, making it a phonetic representation of the language. Example 1 demonstrates how a Persian phrase is written in both Tajik and Farsi.

While proprietary tools such as Google Translate[1] are capable of translating between the two varieties, they have limitations. As translation quality largely depends on the amount of data available, low-resource language pairs like Tajik and Farsi suffer from a lack of large amounts of diverse data (Haddow et al. 2022). Furthermore, as these are two varieties of the same language, machine translation (which does not preserve exact wording) between the two is absolutely fatal to reading poetry (a beloved piece of Persian-speaking culture) and also renders direct quotations of text impossible. On a finer level, this prevents the communication of lexical subtleties.

(1) زبان فارسی

*забони форсӣ*
zaboni forsī

'The Persian language'

Due to their mutual intelligibility in spoken form and the small amount of parallel Tajik-Farsi data, machine transliteration has been proposed as a practical alternative. Early work includes a proof-of-concept system by Megerdoomian & Parvaz (2008) for Tajik to Farsi conversion incorporating a Finite-State Transducer (FST) and lexical lookup. Later, Davis (2012) then introduced a statistical machine transliteration model for transliteration in both directions. Both of these methods lacked corpora made up of sentences with the exact same wording, instead opting for unrelated Tajik and Farsi corpora and a medium-size word list. While functional, such data does not allow for straightforward evaluation metrics on the sentence level (such as sequence accuracy) and cannot capture features that span across several words (such as the Persian Ezafe).

Grapheme to Phoneme Conversion (G2P) is a task that converts letters (grapheme sequence) to their corresponding pronunciations (phoneme sequence) (Yolchuyeva et al. 2019). As a pair of non-phonetic and phonetic representations of the same language, transliteration between Tajik and Farsi resembles G2P. We base our transliteration method on this and treat the task as one of G2P conversion. The main goal of this chapter is to present the current findings of this approach. The preliminary results of our ongoing investigation, drawn from the novel digraphic corpus we created, also provide a benchmark for future work. As a secondary goal, this chapter aims to provide a detailed walkthrough of the many challenges involved in Tajik-Farsi transliteration for others who may attempt similar work in the future.

---

[1] `https://translate.google.com` (Date of Access: 09/30/2023)

The chapter is organized as follows: we first provide an overview of the many differences between Tajik and Farsi (Section 2). Following this, we provide a substantial description of the challenges in transliteration between the two (Section 3) and describe previous (Section 4) and related work (Section 5). We then discuss our parallel dataset (Section 6). Finally, we describe how we trained our model (Section 7) and analyze the results (Section 8) before ending with our concluding remarks (Section 9).

## 2  Tajik and Farsi

### 2.1  Orthographic Differences

Farsi uses the Perso-Arabic script, an impure abjad which often omits vowels. To better reflect Persian phonology, it includes the letters پ /p/, چ /tʃ/, ژ /ʒ/, and گ /g/. As a result of these factors, the script itself is quite consonant-heavy and rather vowel-deficient (Perry 2005).

In contrast, the Tajik-Cyrillic script is an alphabet, meaning that it does not omit any sound and so more accurately reflects Persian phonology. Like Perso-Arabic, it includes several letters that are nonexistent in its source orthography. These letters are Ҳ /h/, Ҷ /dʒ/, Қ /q/, Ғ /ʁ/, Ӯ /ɵ/, Ӣ /i/. As the Tajik-Cyrillic script does not include Arabic-specific characters, ambiguity and homonymy for Arabic loanwords has been maximized (Megerdoomian & Parvaz 2008). This can be seen in the two Farsi words ستر and سطر both pronounced /satr/ and written as сатр in Tajik.

### 2.2  Phonological Differences

While the formal languages have retained mutual intelligibility, they have nonetheless evolved separately for several centuries and noticeably diverged in pronunciation (Perry 2005). For example, Tajik retains the Classical Persian distinction between ʃir "milk" and ʃer "lion" (written as шир and шер, respectively). In Farsi, these are homonyms, both pronounced as /ʃir/ and written as شیر . On the other hand, Farsi distinguishes between پول / pul/ ("money") and پل /pol/ ("bridge") (both written as пул /pul/ in Tajik). The widespread merger of /q/ and /ɣ/ (in Tajik /ʁ/) into [ɣ~ɢ] in many varieties of Farsi is also significant, although the distinction is preserved in both orthographies as ق and غ, in Farsi and in Tajik as Қ and Ғ.

#### 2.2.1  Loanwords

The two varieties have also adopted words from different sources, particularly in the realm of science and technology (Perry 2005). Some of the most common examples include the word for "thank you" (French-derived Farsi "mersi" vs. native Tajik "rahmat") and "potato" (native Farsi "seb-e zamin" vs. Russian-derived Tajik "kartoshka"). Due to recent efforts to remove Russian influence in Tajikistan, many words in the formal Tajik standard are shared with Farsi, but continue to differ in colloquial form, such as "airplane" (Farsi and Tajik "havopaimo" vs. Russian-derived Tajik "samalyot"). Owing to the aforementioned "Persianization" and prevalence of Iranian media, Farsi-specific terms present much less of an issue to Tajiki-speakers than the other way around.

## 3 Challenges in Transliteration

As previously described, Farsi and Tajik diverge in a number of ways which render one-to-one letter conversion largely ineffective. An example transliteration employing such a technique can be seen in Table 1.

Table 1: One-to-one Transliteration of Farsi and Tajik

| Farsi | Farsi Translit. | Tajik | Tajik Translit. |
|---|---|---|---|
| من کتاب را خواندم | mn ktob ro xwondm | ман китобро хондам | man kitobro xondam |

### 3.1 One to Many, Many to One

Owing to the incongruous natures of the two scripts, there exist several challenges in ambiguity when transliterating in either direction. This results in many characters mapping to a single character, and vice versa. This difficulty is pronounced particularly in vowels but extends to consonants as well.

#### 3.1.1 Vowels

The character ا, known as *alef*, can represent several different vowels. If in word medial or terminal position, the *alef* typically maps to the Tajik о. In word start position, it only represents о if written with the *maddah* diacritic, making it آ. Otherwise, it can map to several different vowels. This behavior is demonstrated in Table 2.

Table 2: Examples of ا (*alef*) mapping to various vowels

| Farsi | Farsi Translit. | Tajik | Tajik Translit. | English |
|---|---|---|---|---|
| انجمن | anjmn | анҷуман | anjuman | 'organization' |
| انتخاب | antxob | интихоб | intixob | 'choice' |
| امید | amyd | умед | umed | 'hope' |
| او | aw | ӯ | ü | '(s)he' |
| آهنگ | ohng | оҳанг | ohang | 'song' |
| خاردن | xoridn | хоридан | xoridan | 'to itch' |

The letter و, known as *vav*, can map to the vowels y and ӯ, or the consonant в, as shown in Table 3.

The letter ی, known as *ye*, maps to several different vowels, depending on its position in the word. In addition, in some words of Arabic origin, the character can map to when in word final position. Table 4 provides examples of the character in various positions.

Although not a vowel, the consonant ه, known as *he (do cheshm)*, maps to either the consonant ҳ or to vowels when in word final position. This occurs in a large number of common words, as shown in Table 5.

The character ع, or *ayn*, displays not one, but two separate challenges. The first is that it can map to any vowel (see Table 6), and second is that after the mapped vowel the glottal stop sign ъ is inconsistently written. Moreover, the glottal stop sign also denotes actual glottal stops, and so can be written in words in the absence of *ayn*. However, the erroneous omission or addition of the glottal stop character rarely leads to difficulties in comprehension.

Table 3: Examples of و (*vav*) mapping to vowels and consonants

| Farsi | Farsi Translit. | Tajik | Tajik Translit. | English |
|---|---|---|---|---|
| ولایت | wlayt | вилоят | viloyat | 'province' |
| آورد | owrd | овард | ovard | 'brought' |
| گاو | gaw | гов | gov | 'cow' |
| بود | bwd | буд | bud | 'was' |
| امروز | amrwz | имрӯз | imrüz | 'today' |

Table 4: Examples of ی (*ye*) mappings

| Farsi | Farsi Translit. | Tajik | Tajik Translit. | English |
|---|---|---|---|---|
| یراق | yraq | ярок̧ | yaroq | 'weapon' |
| دریا | drya | дарё | daryo | 'river/sea' |
| چای | çay | чой | çoy | 'tea' |
| ایران | ayran | Эрон | Eron | 'Iran' |
| خیلی | xyly | хеле | xele | 'very' |
| عالی | 'aly | олӣ | olī | 'great' |
| حتی | hty | ҳатто | hatto | 'even' |

The ء, or *hamza*, exhibits similar behavior to *ayn* by mapping to both vowels and the Tajik glottal stop sign. It can be written as a standalone letter, or over *alif*, *vav*, or *ye*. However, this character is often replaced with a *ye* or simply removed from the letter it is written over.

Vowel diacritics are also often unwritten in the Perso-Arabic script, further obfuscating short vowel determination. Without vowel diacritics, the word گرد may represent either гард /gard/ ('dust'), гирд /gird/ ('round'), or гурд /gurd/ ('hero').

### 3.1.2 Consonants

As the Perso-Arabic script retains many redundant consonants from Arabic, some Cyrillic letters each have multiple Perso-Arabic letters equivalents. Table 7 provides an overview of these sounds and the corresponding letters in each script.

Outside of these redundant consonants and the vowels mentioned previously, consonant to consonant mapping between the two scripts is one-to-one and can be considered trivial.

### 3.2 Capitalization

As a modified Cyrillic alphabet, Tajik-Cyrillic capitalizes proper nouns and the beginning of sentences while the Perso-Arabic script does not capitalize at all. When transliterating from Farsi to Tajik, a complete system will need employ a named-entity recognition step, while no additional step is required in the other direction.

Table 5: Examples of ه (*he*) Mapping

| Farsi | Farsi Translit. | Tajik | Tajik Translit. | English |
|---|---|---|---|---|
| به | bh | ба | ba | 'to' |
| که | kh | ки | ki | 'that (conj.)' |
| چه | çh | чи | çi | 'what' |
| قاعده | qa'dh | қоида | qoida | 'rule' |
| سیاه | syah | сиёҳ | siyoh | 'black' |
| ده | dh | даҳ | dah | 'ten' |
| فربه | frbh | фарбеҳ | farbeh | 'fat' |

Table 6: Examples of ع (*ayn*) Mapping mapping

| Farsi | Farsi Translit. | Tajik | Tajik Translit. | English |
|---|---|---|---|---|
| عضو | 'zw | узв | uzw | 'limb' |
| علامت | 'lamt | аломат | alomat | 'sign' |
| فعالیت | f'alyt | фаъолият | fa'oliyat | 'activity' |
| ساعت | sa't | соат | soat | 'hour' |
| تاریخ | taryx | таърих/торих | ta'rix/torix | 'history' |
| قرآن | qron | Қуръон | Qur'on | 'Quran' |

## 3.3 Persian Ezafe/Izofat Detection

The Persian Ezafe (known as izofat in Tajik) is a Persian grammatical feature that links an modifier to a preceding head noun (or preceding modifier) through the unstressed affix /-i/ (or /-yi/) in Tajik, and /-e/ (or /-ye/) in Farsi (Perry 2005). Example 2 demonstrates its usage to link an adjective to a head noun, and Example 3 demonstrates its usage to denote possession.

This affix manifests itself in different ways in both scripts. In accordance with its phonetic nature, the Tajik standard always writes the Ezafe as an -и attached to the previous word. When it is added to a word that ends with -ӣ, the accent is dropped and it becomes -ии. However, when it is added to a word with -й, either the accent is dropped and it becomes -и or it is preserved.

In contrast, the Perso-Arabic script almost always omits the Ezafe, so the reader must infer its location from the surrounding context. Typically, the Ezafe is only always written as a ی when added to the plural marker /-ho/ها.This behavior can be seen in Examples 4 and 5. Otherwise, if it is absolutely needed to disambiguate a phrase, it is written as a a diacritic at the end of the head noun.

(2) خانه زیبا
хонаи зебо
xona-i zebo
house-EZ beautiful

'The beautiful house'

Table 7: One to many mappings of consonants from Tajik to Farsi

| Phoneme | Tajik | Farsi |
|---|---|---|
| /z/ | з | ز |
|  |  | ذ |
|  |  | ض |
|  |  | ظ |
| /s/ | с | س |
|  |  | ص |
|  |  | ث |
| /t/ | т | ت |
|  |  | ط |
| /h/ | ҳ | ح |
|  |  | ه |

(3) خانه مرد
   хонаи мард
   xona-i mard
   house-EZ man

   'The man's house'

(4) خانه‌های مرد
   хонаҳои мард
   xona-ho-i mard
   house-PL-EZ man

   'The man's houses'

(5) خانه‌های زیبا مرد
   хонаҳои зебои мард
   xona-ho-i zebo-i mard
   house-PL-EZ beautiful-EZ man

   'The man's beautiful houses'

The Ezafe has the potential to drastically change a given sentence's phrasal boundaries, and subsequently, its meaning (Doostmohammadi et al. 2020). As such, a given sentence in Farsi can be written one way but interpreted in several, as demonstrated in Examples 6 and 7 (with explicitly marked Ezafe).

(6) در پیِ اتفاقاتِ دیروز او استعفا داد
   Дар пайи иттифоқоти дирӯз ӯ истеъфо дод dar pay-i ittifoqot-i dirüz ü iste'fo dod
Following-EZ events-EZ yesterday (s)he resign did

'Following yesterday's events, he/she resigned'

(7) در پیِ اتفاقات دیروز او استعفا داد
Дар пайи иттифоқот дирӯз ӯ истеъфо дод
dar pay-i ittifoqot dirüz ü iste'fo dod
Following-EZ events yesterday (s)he resign did

'Following the events, (s)he resigned yesterday'

While the word-final -и can simply be ignored when converting from Tajik to Farsi, the opposite direction is another matter. When converting from Farsi to Tajik, a complete system must infer the Ezafe's location, which can only be determined from semantic - rather than orthographic - context. Unfortunately, despite the dearth of literature on the Ezafe, to our knowledge there exists only one publicly available Ezafe detection tool by Etezadi et al. (2022). Moreover, its Ezafe detection accuracy has not been measured and verified.

## 3.4 The Conjunction "*va*"

The conjunction "and" also presents a unique set of problems. In Tajik, its two equivalents are the independent ва /va/ and the enclitic -у/-u/ (Perry 2005). However, in Farsi both forms are written as و. There is no way to distinguish between the two possibilities when transliterating from Farsi to Tajik. An exception to this rule may be poetry, where the surrounding context would guide a reader's pronunciation. Examples 8 and 9 demonstrate these different forms with a single Farsi phrase from our dataset.

This causes several issues for transliteration from Farsi to Tajik. The first is that it presents our models with two equally-valid transliterations of the same word. This could act as "noise" for our model, and result in a completely invalid transliteration. It also complicates efforts to evaluate the word-level and character-level accuracy of our model.

(8) تاجیک و ایرانی و افغان
тоҷик ва эронӣ ва афғон
tojik va eronī va afghon
Tajik and Iranian and Afghan

(9) تاجیک و ایرانی و افغان
тоҷику эрониву афғон
tojik-u eroni-vu afghon
Tajik-AND Iranian-AND Afghan

## 3.5 Zero-width Non-joiner

The Zero-width Non-joiner (ZWNJ) is a very common yet invisible (non-printing) character used solely when writing Farsi with a keyboard. Placing it between two Farsi characters causes them to be written as if they were separated, meaning in their initial and final forms, respectively. In informal texts, it is common to see the ZWNJ either simply removed or replaced with a space. As a general rule, words written with a ZWNJ in Farsi are also written as single words in Tajik.

There are several situations in which the ZWNJ is used: affixes, verbal auxiliaries, the Persian object marker *ro*, the present tense conjunction, and compound nouns.

Table 8 and Table 9 demonstrate the ZWNJ with examples for the present tense conjunction and the following affixes: the plural marker, the Persian object marker "ro", and the verbal auxiliary.

Table 8: Different Ways to Separate Affixes

| Farsi with ZWNJ | Farsi without ZWNJ | Farsi with Space |
| --- | --- | --- |
| موضوع‌را | موضوعرا | موضوع را |
| خانه‌ها | خانهها | خانه ها |
| می‌خواهم | میخواهم | خواهم می |
| کتاب‌فروش | کتابفروش | فروش کتاب |

Table 9: Separated Affixes in Farsi

| Farsi | Farsi Translit. | Tajik | Tajik Translit. | English |
| --- | --- | --- | --- | --- |
| کرده است | krdh ast | кардааст | kardaast | 'has done' |
| خانه‌ها | xanh ha | хонаҳо | xonaho | 'houses' |
| موضوع‌را | mwzw' ra | мавзӯъро | mavzü'ro | 'the subject' |
| می‌خواهم | myxwahm | мехоҳам | mexoham | 'I want' |
| کتاب‌فروش | ktabfrwš | китобфурӯш | kitobfurüš | 'book seller' |

## 3.6 The Glottal Stop 'ъ'

This character ъ represents the glottal stop and typically corresponds to an instance of Farsi *'ayn* ع. However, it often presents inconsistencies, occurring in the absence of *'ayn* or disappearing when *'ayn* is present. The most common example of this is the word соат/ساعت meaning 'hour'. An incorrect transliteration of this word would result in соъат instead. This character can even be duplicated, such as in the word фаъолият which is sometimes written as фаъъолият. Although the character is extremely common, omission of this character does not hinder comprehension on a sentence level.

### 3.7 The Dash Character in Tajik

## 4 Previous Work

### 4.1 Finite-State Transducer

Megerdoomian & Parvaz (2008) developed the first-ever system designed to tackle Tajik to Farsi transliteration, creating a proof-of-concept system based on a finite-state transducer (FST). The FST converts Tajik to Farsi, overgenerating a number of possible transliterations for the input data based on a number of given contextual rules. All results are then subject to morphological analysis and a dictionary lookup. If the system finds morphological match or lexical match, then that match is used (in that order of testing). If no match is found, it selects the most-likely form using "rules of thumb" based letter frequencies. Their system achieved 89.8% accuracy in transliterating a document from Tajik to Farsi, and generated 6.27 alternative spellings for the average token.

### 4.2 Statistical Machine Transliteration

Davis (2012) introduced a phrase-based statistical machine transliteration model for Tajik-Farsi transliteration. In this model, the transliteration model translates at the word-level using character-based n-grams, rather than at the word-level using word-based n-grams. This work created character-level language models based on Tajik and Farsi corpora made from online news sources and a training corpus of 3,503 pairs. Due to a lack of parallel corpora, evaluation of model's accuracy was conducted using two tasks: part of speech (POS) tagging and machine translation. For part of speech tagging, a Tajik POS tagger was trained on transliterated text and its accuracy was compared to a subset of the corpus to the reported accuracy of other Farsi POS taggers on the unrelated Bijankhan corpus. The SMT model achieved an accuracy of 92.52%. In the machine translation task, he transliterated Tajik to Farsi, and then translated the resulting Farsi into English using Google Translate. The system achieved a BLEU score of 0.2349. As this BLEU score was ≈90% within Google's BLEU score for Farsi to English, the system's raw accuracy was approximated at ≈90%. However, as both BLEU score and PoS tagger accuracy are not meant to be compared across datasets without recalculation, we consider these results invalid and do not reference them.

## 5 Related Work

As stated earlier, machine transliteration is typically used for the transcription of named entities, rather than translation to a highly related dialect. However, that does not mean that Tajik and Farsi are in a singular situation. Several other languages present similar levels of intelligibility and digraphia, and machine transliteration has been investigated in these cases as well. Ahmadi (2019) presented a rule-based approach for Sorani Kurdish transliteration between an Arabic-based and a Latin-based orthography. Talafha et al. (2021) describe a model based on Recurrent Neural Networks (RNN) for transliteration between Jordanian Arabic written in the Arabic script and in a non-standard romanization known as Arabizi. For the extinct Ottoman Turkish, Jaf & Kayhan (2021) explored rule and lexicon-based transliteration to aid documentation efforts. Durrani et al. (2010) trained conditional and joint probability models for Hindi-Urdu, also two closely-related varieties of one macrolanguage. Following recent orthography reforms, Salaev et al. (2022) explored rule-based transliteration for Uzbek.

## 6  Dataset

### 6.1  Dataset Compilation

To effectively train and evaluate our model, we require Tajik-Farsi parallel data aligned on a word-to-word basis. Unfortunately, due to lexical and grammatical differences and translator bias, parallel texts sourced from organizations such as the United Nations or Jehovah's Witnesses generally contain extremely divergent sentence pairs. This renders usage of such texts inappropriate for our purposes. To amass as much data as possible, we collected data from various sources on the Internet. Our efforts created the very first digraphic Tajik-Farsi corpus (Merchant & Tang 2024). On top of this data, we include a digraphic dataset released by another group after our own. Tables 10 and 11 provide comparison of the two corpora. We note that the second corpus consists of both a dictionary and poetry collection, so we do not include sentence statistics. To our knowledge, both datasets have not undergone in-depth analysis from native speakers and therefore may occasionally contain incorrect correspondences. We note that creation of digraphic dataset checked by native speakers presents another avenue for research.

We determined combined usage of these two datasets to be sufficient for our purposes based on related work in transliteration. In the case of Tajik and Farsi, Davis (2012) used a list of only 3,503 words. In the NEWS 2018 Shared Task on Machine Transliteration (Chen et al. 2018), the submitted neural models achieved F-scores of almost 80% or more on language pairs with datasets ranging in same from 6,000 to 41,318 individual names per language pair. Taking this into account we judge our dataset to be more than adequate, especially considering the fact that our model operates at the character level.

Table 10: Our Corpus Statistics *Note we did not consider the whitespace character

| Statistics | Farsi | Tajik |
| --- | --- | --- |
| # of sentences | 2,813 | 2,813 |
| # of word tokens | 43,846 | 42,226 |
| # of characters* | 186,414 | 222,986 |
| Avg. # of tokens in a sentence | 15.59 | 15.01 |
| Avg. # of characters* in a sentence | 66.27 | 79.27 |

Table 11: Stibiumghost Corpus Statistics *Note we did not consider the whitespace character

| Statistics | Farsi | Tajik |
| --- | --- | --- |
| # of word tokens | 403,829 | 392,924 |
| # of word tokens (excl. dictionary) | 360,294 | 349,389 |
| # of characters* | 1,555,580 | 1,862,938 |
| # of characters* (excl. dictionary) | 1,258,628 | 1,533,549 |

### 6.1.1 Online Blogs and BBC Articles

After an extensive search, we found two main sources of parallel data online: blogs and British Broadcasting Corporation (BBC) News. The two blogs we found were written by Persian-speakers who knew both scripts and dealt with a wide variety of topics that include poetry and politics[2][3]. To filter our posts written in only script or other languages (mainly Russian), we opted to manually collect these data rather than use an automatic website scraping tool. We were also able to find 23 BBC News articles written in both scripts[4][5]. These articles almost exclusively deal with politics, and exhibited a similar degree of word-to-word alignment. Due to the small number of articles, we elected to collect these by hand as well. Our corpus contains 2,813 Tajik-Farsi sentences with average Tajik and Farsi sentence lengths of 15.59 and 15.01 words respectively, as shown in Table 10 (Merchant & Tang 2024). We note that our our corpus will eventually be made freely available to the broader linguistics community.

### 6.1.2 Dictionary and Poetry

Several months after creating our dataset, another group released their own set of digraphic Persian data on GitHub for the purpose of Tajik to Farsi transliteration (stibiumghost 2022). This includes parallel news texts, a Classical Persian poetry collection, and a Farsi-Tajik dictionary. This data is publicly available on the project's GitHub repository. The dictionary contains 43,525 entries, and the rest of the corpus consists of 37,940 Tajik-Farsi sentences with average Tajik and Farsi sentence lengths of 49 and 42.91 words respectively. The combined statistics has been shown in Table 11.

## 6.2 Data Processing

### 6.2.1 Alignment and Normalization

As the corpus that we compiled was not aligned on a sentence-to-sentence basis, we utilized GaChalign, a Python implementation of the Gale-Church alignment algorithm, to do so (Tan & Bond 2014). Afterwards, we normalize the data by stripping it of any and all characters that are not Perso-Arabic or Tajik-Cyrillic letters, including Arabic diacritics, numbers, and punctuation. We also strip the hyphen character from our Tajik data, which is often used to show that a certain word is made up of two words. This is typically seen in poetry, for example the single word к-аз, made up of ки (which) and аз (from). Preliminary testing showed that this character only provided additional noise to our data, so we deemed it unnecessary.

### 6.2.2 Separated Affix Normalization

As described earlier, certain affixes in Farsi can be written in several different ways. As the authors of our data incorporate both the ZWNJ and a single space to separate these affixes, we opted to normalize this data. To do so, we connected all the affixes we could find (using a rule-based approach) to the preceding stem using the ZWNJ. We provide the ZWNJ to our model as an an acceptable Farsi grapheme.

---

[2]https://dariussthoughtland.blogspot.com/
[3]https://jaamjam.blogspot.com/
[4]https://www.bbc.com/tajik
[5]https://www.bbc.com/persian

## 7 System Description

### 7.1 Transformer-based G2P Conversion

The transformer architecture has become an increasingly popular and successful choice for sequence-to-sequence conversion in recent years, rivaling the previously state-of-the-art recurrent neural networks (RNNs) with less data and training time (Wu et al. 2021). Previously applied mainly to larger tasks, such as machine translation, question answering, and summarization, more and more research has begun applying it to character and sub-word level tasks, such as G2P conversion, transliteration, and morphological inflection. Here too, the model has replicated its initial success [6]. We train the model in both directions, Farsi to Tajik and Tajik to Farsi.

We apply a encoder-based seq2seq model based on that proposed by Yolchuyeva et al. (2019) using the DeepPhonemizer tool from Schäfer et al. (2023). The transformer architecture consists of 4 encoder and decoder layers, with four parallel attention layers. The size of the embeddings is 256 and the dimension of the feed-forward neural network inner layer is 1024. Our code[7] and data[8] are both available available to access online.

### 7.2 Model Training

To prepare our dataset, we utilize ten-fold cross validation, ensuring that for each model interation, we use a unique (and randomized) split. Our overall dataset is split into 80% training, 10% dev, and 10% test sets, with this split being applied to each individual source of data (dictionaries, blogs, poetry etc.)

We trained our model with an initial learning rate of 0.001 and multiplied the value by 0.5 if performance on the validation set did not improve for 10 epochs. We also implement a warmup phase of 5 epochs, and set the dropout rate for both the encoder and decoder to 0.1 to combat overfitting (Srivastava et al. 2014). Due to hardware limitations, batch size is set to 16. Training hyperparameters are not changed when training in either direction. With these settings, we found that training our model past 100 epochs did not lead to significant differences in accuracy.

## 8 Evaluation and Results

### 8.1 Metrics

As this task blends translation with transliteration, we report several evaluation metrics to evaluate our model from different viewpoints [9]. As previous efforts presented indirect metrics due to a lack of parallel data, the results provided in Table 12 also set a baseline for future efforts (including our own).

---

[6] We are aware that LLMs could also be utilized for this task, but elected to only use transformers due to the abundance of literature applying transformer models to transliteration and the lack thereof for LLMs.

[7] https://github.com/merchantrayyan/NACIL_ParsTranslate/tree/main

[8] https://github.com/merchantrayyan/ParsText

[9] As we only present one model, we do not present a trajectory of learning over epochs. However, in future, different hyperparameters should be experimented with and the results of these across epochs could then be compared.

*Character F-score (chrF)* is based on the intuition that more accurate translations will tend to have characters that appear in a human's translation. To do so, it compares the number of character n-gram overlaps between the system output and the reference transliteration. While the base implementation only looks at character n-grams (and is therefore slightly optimistic), recent research from Popović (2017) has shown that a modified version (chrF++), which takes unigrams and bigrams into account, better reflects human judgement. As such, we provide chrF++ as our main metric.

*Sequence Accuracy* is the percentage of model outputs that are exactly the same as their expected counterpart. Due to differences in word segmentation between Farsi and Tajik, this is calculated by removing all spaces from both the reference transliteration and the system output and comparing the two resulting sequences.

*Relaxed Sequence Accuracy (RSA)* denotes the total number of correct sequences combined with the number of sequences one edit distance away and vice versa. A single edit distance is defined as a single addition, subtraction, or substitution to the predicted sequence that would make it exactly the same as the reference sequence.

*Average Levenshtein Distance (Avg. LD)* is realized as the average Levenshtein (or edit) Distance between the expected output sequence and the model prediction.

*Levenshtein Distance Ratio (LD Ratio)* is taken as the ratio of the Levenshtein Distance to the length of max(expectedoutput, modeloutput). Given an expected sequence and a reference sequence, a NPER of 0.0 means that the sequences are exactly identical, while a NPER of 1.0 means that they have nothing in common.

*F-score* is deemed a more appropriate metric for character-level evaluation due to the fact that raw accuracy would not accurately reflect less-common characters. We created a confusion matrix by aligning reference translations to our system output, and then calculate the overall F-score as well as the F-score for consonants and vowels. This data is presented in Tables 13, 14, and 15.

## 8.2 Results

Interestingly, our results revealed that our model performed significantly better on Tajik-to-Farsi transliteration than Farsi-to-Tajik across all metrics, seen in Table 12. Although the Sequence Accuracy between Farsi-to-Tajik and Tajik-to-Farsi (with ZWNJ) are within 0.4% of each other, the chrF++ scores of 58.70 and 65.43 show a greater difference in performance. Interestingly, omitting the ZWNJ drastically increases Sequence Accuracy for Tajik-to-Farsi from 34.37% to 50.46%, chrF++ from 65.43 to 74.20, and decreases the Levenshtein Distance Ratio (L.D. Ratio) from 0.09 to 0.05. As the ZWNJ is an invisible character, we consider these results to be much more representative of our model's performance in this direction.

The stark contrast in these directions suggests that Farsi-to-Tajik transliteration poses a much greater challenge to our model than Tajik-to-Farsi transliteration. Comparison of the F1 scores shown in Tables 13 and 14 reveals vowel prediction to be the primary factor, with a weighted F1 score of 0.85 from Farsi-to-Tajik lagging behind the 0.99 achieved by Tajik-to-Farsi. A contributing factor to this may have been the surprising inability of our model to correctly transliterate the character *he* when in word-final position, despite the prevalence of the prepositions کَ and بِ (ки and ба) as seen in Examples 10 and 11. Beyond vowel prediction, our model also failed to predict semantically-important features such as the conjunction "*va*" and the Ezafe, as seen in Examples 10, 11, and 12.

Although our model performed better on Tajik-to-Farsi transliteration, several issues are stillclear. Most notably, prediction of the ZWNJ proves to be more difficult than anticipated, with its omission causing significant improvements across all metrics. Since the ZWNJ does not impact comprehension, our model still outputs quite understandable transliterations, as seen in Examples 13, 14, and 15. Of course, some small errors remain, as demonstrated by the Macro F1 score of 0.95 in 15, and the unsuccessful deletion of the vowel "*vav*" in Example 15. The Macro F1 score in Table 13 in particular suggests that mapping to the redundant Farsi consonants may have caused a small issue. Although such incorrect mapping has greater semantic implications than the ZWNJ, it would cause only minimal confusion for a native speaker.

For all examples, the first line is the input sequence, the second line is the reference (correct) transliteration, and the final line is our system output. Incorrect and omitted letters are indicated in bold.

Table 12: Evaluation metrics for model performance in both directions

| Model | chrF++ | Seq. Acc. | RSA+1 | RSA+2 | Avg. L.D. | L.D. Ratio |
|---|---|---|---|---|---|---|
| Farsi→Tajik | 58.70 | 33.99% | 46.55% | 52.39% | 3.79 | 0.11 |
| Tajik→Farsi (ZWNJ) | 65.43 | 34.37% | 56.78% | 69.68% | 2.12 | 0.09 |
| Tajik→Farsi (No ZWNJ) | 74.20 | 50.46% | 70.59% | 82.86% | 1.38 | 0.05 |

Table 13: F1 scores for predicted consonants in both directions (without ZWNJ)

| | Macro F1 | Micro F1 | Weighted F1 |
|---|---|---|---|
| Farsi→Tajik | 0.97 | 0.98 | 0.98 |
| Tajik→Farsi | 0.94 | 0.99 | 0.98 |

Table 14: F1 scores for predicted vowels in both directions (without ZWNJ)

| | Macro F1 | Micro F1 | Weighted F1 |
|---|---|---|---|
| Farsi→Tajik | 0.80 | 0.86 | 0.85 |
| Tajik→Farsi | 0.97 | 0.99 | 0.99 |

(10) نگو که چشم مرا مونس نگاه نبود
нагӯ ки чашм**и** маро мӯнис**и** нигоҳ набуд

наг**у** к**аҳ** чашм м**у**ро м**у**нис ниг**аҳ** набуд

(11) این باز مارا به بحث خط و زبان می‌آرد
ин боз моро ба баҳс**и** хат**ту** забон меорад

ин боз моро б**еҳ** баҳс хат **в** забон меорад

Table 15: F1 scores for all predicted characters in both directions (without ZWNJ)

|  | Macro F1 | Micro F1 | Weighted F1 |
|---|---|---|---|
| Farsi→Tajik | 0.91 | 0.93 | 0.90 |
| Tajik→Farsi | 0.95 | 0.99 | 0.98 |

(12) که ز خرمنهای خوش اعمی بود
می‌کشد آن دانه‌را با حرص و بیم
ки зи хирманҳои хуш аъмо бу**ва**д
мекашад он донаро бо ҳирс**у** бим

к**аҳ** з**е** хирманҳ**ӣ** хв**а**ш аъмо буд
м**ай**кашад он донаро б**ӣ** ҳирс **в** бим

(13) забони давлатии Тоҷикистон забони тоҷикӣ аст
زبان دولتی تاجیکستان زبان تاجیکی است

زبان دولتی تاجیکستان زبان تاجیکی است

(14) бидонем кистему аз куҷо меоем
بدانیم کیستیم و از کجا می‌آییم

بدانیم کیستیمو از کجا میایم

(15) Гуфт Исрофилро Яздони мо
Ки бирав зон хок пур кун каф биё
گفت اسرافیل‌را یزدان ما
که برو زان خاک پر کن کف بیا

گفت **اسرافیلرا** یزدان ما
که برو زان خاک **پور کون** کف بیا

## 9 Conclusion and Future Work

In this paper, we explored the challenges of Tajik-Farsi transliteration and presented a transformer-based transliteration approach as an alternative to machine translation. Our results prove our approach to be viable, with our model achieving strong chrF++ scores of 58.70 (Farsi-to-Tajik), 65.43 (Tajik-to-Farsi), and 74.20 (Tajik-to-Farsi without the ZWNJ) on novel digraphic datasets. Our usage of digraphic texts by native speakers, as opposed to word lists or non-corresponding texts, allowed for realistic evaluation of model performance inter-word features such as the conjunction "*va*", the ZWNJ, and the Ezafe.

Further analysis revealed our model encountered significant issues in Farsi-to-Tajik transliteration. In this direction, our model achieved lower results across all metrics, most notably a weighted F1 score of 0.85 for vowel prediction. Its inability to detect the Ezafe also negatively impacted sentence meaning. Transliteration in the opposite direction proved to be much easier, with a 16.09% increase in Sequence Accuracy when the semantically-trivial ZWNJ was omitted. Overall, our model generally outputs quite understandable Tajik-to-Farsi transliterations. In both directions, our model's results set a strong baseline across a variety of metrics, thus allowing for comparison with future efforts and highlighting the key challenges Tajik-Farsi transliteration presents.

The success of this approach also suggests that it can be applied to other cases of digraphia with non-phonetic and phonetic written standards, including but not limited to: Arabic (Talafha et al. 2021), Hindi-Urdu (Durrani et al. 2010), Kurdish (Ahmadi 2019), Azeri (Zohrabi et al. 2023), Konkani (Rajan 2014) and even Ottoman Turkish (Jaf & Kayhan 2021). In the case of Hindi-Urdu and Ottoman Turkish, both of these languages borrow heavily from Persian to the point where even Ezafe detection would be necessary.

Our future work will focus on hyperparameter tuning, Ezafe detection, and further data acquisition to increase model performance. In particular, our Ezafe detection and additional data from different domains will greatly improve the versality of our model. We hope that our efforts contribute to the development of a practical Tajik-Farsi transliteration tool, further connecting Tajikistan with the broader Persian-speaking world.

**Funding information**

The work was partly performed during a research visit by Rayyan Merchant funded by the Summer Undergraduate International Research Program, University of Florida.

**CRediT authorship contribution statement**

We follow the CRediT taxonomy[10].
 Conceptualization: RM; Data curation: RM; Formal Analysis: RM, AKR, KT; Funding acquisition: RM, Investigation: RM; Methodology: RM, AKR, KT; Supervision: AKR, KT; Visualization: RM; Writing: RM; and Writing – review & editing: RM, AKR, KT.

**Acknowledgements**

We would like to thank the audience at the Third North American Conference in Iranian Linguistics (NACIL 3) for their feedback and comments on our work. Significant thanks are due to Dr. Christopher Geissler, Anna Stein, Anh Kim Nguyen, Julika Weber, Lara Rüter, Nina Stratmann, and Ann-Sophie Haan for their support, advice, and assistance throughout this project. All remaining errors are our own.

---

[10]https://credit.niso.org/

**Abbreviations**

FST Finite-State Transducer
POS Part of Speech
BBC British Broadcasting Corporation
G2P Grapheme-to-Phoneme
TTS Text-to-Speech
RNN Recurrent Neural Network
ZWNJ Zero-width Non-joiner